\documentclass[smallcondensed]{svjour3}
\smartqed  

\usepackage{graphicx}
\usepackage{booktabs} 
\usepackage{tabularx}
\usepackage{multirow} 
\usepackage{balance}
\usepackage{color}
\usepackage{makecell}
\usepackage{hyperref}
\usepackage{amsmath}

\newcommand{\new}[1]{{\color{black}#1}}

\newcommand{\q}[1]{\lq\lq{}{}#1\rq\rq{}{}}

\begin{document}

\title{Exploiting stance hierarchies for cost-sensitive stance detection of Web documents}

\author{
Arjun Roy \and 
Pavlos Fafalios \and 
Asif Ekbal \and 
Xiaofei Zhu \and
Stefan Dietze
}


\institute{A. Roy \at
           L3S Research Center, Leibniz University of Hannover,
           Hannover, Germany\\
           \email{roy@L3S.de}         
           \and
           P. Fafalios \at
           Institute of Computer Science, FORTH-ICS,
           Heraklion, Greece\\
          \email{fafalios@ics.forth.gr}
          \and
           A. Ekbal \at
           Dept. of CSE, IIT Patna,
           Patna, India\\
          \email{asif@iitp.ac.in}  
          \and
           X. Zhu \at
           Chongqing University of Technology,
           Chongqing, China\\
          \email{zxf@cqut.edu.cn}  
          \and
           S. Dietze \at
           GESIS - Leibniz Institute for the Social Sciences,
           Cologne \& Heinrich-Heine-University Düsseldorf, Germany\\
          \email{stefan.dietze@GESIS.org}  
}

\date{\phantom{1}}

\maketitle

\begin{abstract}

Fact checking is an essential challenge when combating fake news. Identifying documents that agree or disagree with a particular statement (claim) is a core task in this process. In this context, stance detection aims at identifying the position (stance) of a document towards a claim. 
Most approaches address this task through classification models that do not consider the highly imbalanced class distribution. 
Therefore, they are particularly ineffective in detecting the minority classes (for instance, \lq{}disagree\rq{}), even though such instances are crucial for tasks such as fact-checking by providing evidence for detecting false claims. 
In this paper, we exploit the hierarchical nature of stance classes which allows us to propose a modular pipeline of cascading binary classifiers, enabling performance tuning on a per step and class basis. We implement our approach through a combination of neural and traditional classification models that highlight the misclassification costs of minority classes. Evaluation results demonstrate state-of-the-art performance of our approach and its ability to significantly improve the classification performance of the important \lq{}disagree\rq{} class.

\end{abstract}


\section{Introduction}

Spread of fake news and false claims have become ubiquitous due to the widespread use and network effects facilitated by social online platforms \cite{lazer2018science}. A recent study has shown that false claims are re-tweeted faster, further, and for longer than true claims on twitter \cite{vosoughi2018spread}. Another study found that the top 20 fake news stories about the 2016 U.S. presidential election received more engagement on Facebook than the top 20 election stories from the 19 major media outlets \cite{chang2016fake}. These findings demonstrate the significance and scale of the fake news problem and the potential effects it can have on contemporary society \cite{allcott2017social}. 

In this context, \textit{fact-checking} is the task of determining the veracity of an assertion or statement (a \textit{claim}) \cite{popat2016credibility,wu2014toward,hassan2015detecting}. 
Identifying the particular semantics of relation between a  (Web) document and a given statement is an essential task, which is widely referred to as \textit{stance detection} \cite{kucuk_stance_2020}. More precisely, stance detection aims at identifying the stance (perspective) of a document towards a claim, namely whether the document \textit{agrees} with the claim, \textit{disagrees} with the claim, \textit{discusses} about the claim without taking a stance, or is entirely \textit{unrelated} to the claim.
However, the real-world distribution of the aforementioned classes is highly imbalanced, where in particular instances of classes of crucial importance (\textit{related}, \textit{disagree}) are strongly underrepresented. This is reflected by state-of-the-art benchmark datasets such as the \textit{FNC-I} dataset \cite{pomerleau2017fake}, where the \textit{disagree} class corresponds to less than 3\% of the instances. 

Existing approaches that try to cope with this problem are ineffective in detecting instances of minority classes. For instance, whereas the overall performance of state-of-the-art systems \cite{baird2017talos,hanselowski2017description,riedel2017simple,lyon:18,f1_retrospec:19,zhang2019stances,masood2018fake} ranges between 58\% and 61\% (F1 macro-average), the performance on the \textit{disagree} class ranges between 3\% and 18\% only. However, this class is of key importance in fact-checking since it enables detecting documents that provide evidence for invalidating false claims.
It is worth noting that there also have been  efforts to deal with a similar problem by simply introducing a  class-wise penalty into the loss function \cite{lin2017focal,Veropoulos99controllingthe} , thereby limiting its capacity to  solve the class imbalance problem.

\begin{figure}
	\fbox{\includegraphics[width=2.5in]{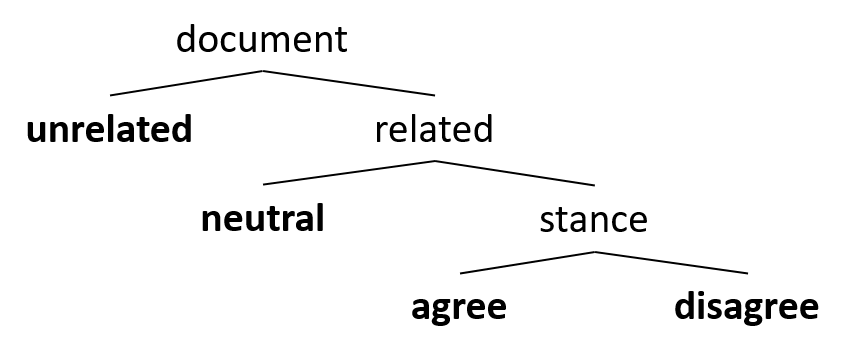}}
	\caption{Document stance hierarchy.}
	\label{fig:tree}
\end{figure}

In this paper, we exploit the hierarchical nature of stance classes (Figure \ref{fig:tree}) by introducing a classifier cascade of three binary classification models, where individual tuning at each step enables better consideration of misclassification costs of minority classes. This is aimed at improving the classification performance of the important but underrepresented \textit{disagree} class without negatively affecting the performance of other classes. 
To this end, we propose a three-stage pipeline architecture that treats the 4-class classification problem as three different binary classification (sub-)tasks of increasing difficulty.
The first stage aims at detecting the documents related to the claim (\textit{relevance classification}), the second stage classifies only related documents and focuses on detecting those that take a stance towards the claim (\textit{neutral/stance classification}), and the third stage classifies documents taking a stance as agreeing or disagreeing to the claim (\textit{agree/disagree classification}). 
This modular and step-wise approach offers the flexibility to consider different classifiers and features in each different stage of the pipeline depending on the context and corpus at hand, thus enabling to optimise performance on a per stage and class basis.

We provide an implementation\footnote{\url{https://github.com/arjunroyihrpa/Stance-Hierarchies}} of our pipeline architecture, called \textit{L3S (Learning in 3 Steps)}, using a combination of neural and traditional classification models and by introducing cost-sensitive measures to reflect the importance of minority classes. Evaluation results following the established stance detection benchmark\cite{pomerleau2017fake} show that our approach achieves the state-of-the-art performance on the general stance classification problem, slightly outperforming all the existing methods by one percentage point (macro-average F1 score). Most importantly, we significantly improve the F1 score of the \textit{disagree} class by 28\% compared to the state-of-the-art.

The rest of the paper is organized as follows: 
Section \ref{sec:modeling} formulates the problem and provides an overview of our pipeline approach.
Section \ref{sec:pipeline} describes supervised models for each stage of the pipeline. 
Section \ref{sec:eval} reports evaluation results. 
Section \ref{sec:rw} presents related works.
Finally, Section \ref{sec:conclusion} concludes the paper.

\section{Problem Description and Approach Overview}
\label{sec:modeling}

Given a textual claim $c$ (e.g., \textit{\q{KFC restaurants in Colorado will start selling marijuana}}) and a document $d$ (e.g., an article), stance classification aims at classifying the stance of $d$ towards $c$ to one of the following four categories (classes): 

\begin{itemize}
\item \textbf{Unrelated}: the document is not related to the claim. 
\item \textbf{Neutral}: the document discusses about the claim but it does not take a stance towards its validity. 
\item \textbf{Agree}: the document agrees with the claim. 
\item \textbf{Disagree}: the document disagrees with the claim.
\end{itemize}

These four classes can be structured in a tree-like hierarchy as shown in Fig. \ref{fig:tree}. 
At first, a document can be either \textit{unrelated} or \textit{related} to the claim. 
Then, a document that is related to the claim can either be \textit{neutral} to the claim or take a \textit{stance}. Finally, a document that takes a stance can either \textit{agree} or \textit{disagree} with the claim. The leaves of the tree are the four classes. 
Considering this structure, we can now model the stance classification problem as a three-stage classifier cascade consisting of three connected sub-tasks (or \textit{stages}):

\begin{itemize}
\item   \textbf{Stage 1 (relevance classification)}: identify if a document is related to the claim or not.
\item   \textbf{Stage 2 (neutral/stance classification)}: identify if a document classified as \textit{related} from stage 1 is neutral to the claim or takes a stance.
\item   \textbf{Stage 3 (agree/disagree classification)}: identify if a document classified as \textit{stance} from stage 2 agrees or disagrees with the claim.
\end{itemize}

\begin{figure}
	\includegraphics[width=4.5in]{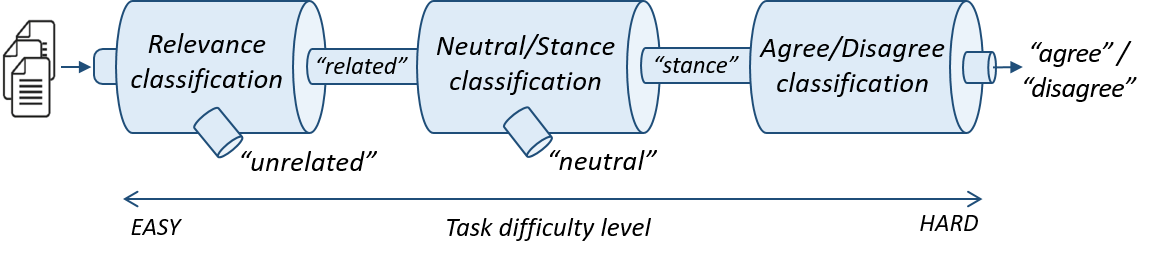}
	\caption{Pipeline for document stance classification and difficulty level of each stage.}
	\label{fig:pipeline}
\end{figure}

Figure \ref{fig:pipeline} depicts the proposed pipeline architecture. 
Each stage of the pipeline can now be modeled as a separate binary classification problem.  We hypothesize that relevance classification (stage 1) can filter out documents unrelated to the claim which can facilitate the neutral/stance classification task (stage 2). Likewise, knowing that a document takes a stance towards the claim can facilitate the agree/disagree classification task (stage 3). 
A weakness of such a pipeline approach is that errors can propagate from one stage to the other, thus errors in earlier stages negatively affect the later stages. It is to be noted though that, in general, relevance classification (stage 1) is a much easier task than neutral/stance classification (stage 2), which in turn is considered easier than agree/disagree classification (stage 3). Our experimental evaluation validates this hypothesis (more in Sect. \ref{sec:eval}). 

\section{Pipeline Implementation}
\label{sec:pipeline}

In this section we describe our classification models for each stage of the pipeline, each being implemented through a supervised model tailored towards the specific classification problem at hand \new{and using features that reflect the syntactical and semantic similarity between a claim and a document.}

\subsection{Stage 1: Relevance Classification}
\label{stage1}

Existing approaches on separating the \textit{related} from the \textit{unrelated} instances have already achieved a high accuracy ($>$95\%) \cite{f1_retrospec:19}. Such models usually make use of hand-crafted features that aim at reflecting the text similarity between the claim and the document.
Since our focus is on the important \textit{disagree} class (which is part of \textit{related} in this stage), we seek a model that penalises misclassifications of this class. To this end, we experimented with a variety of different classifiers, inspired by previous works that perform well on the same problem, including:  
support vector machine (SVM) with and without class-wise penalty, gradient boosting trees, AdaBoost, decision trees, random forest with and without class-wise penalty, and convolutional neural networks (CNN). 

To train the classifiers, we use the below set of features, selected through extensive feature analysis. 
The first four features are used in the baseline model provided by FNC-I \cite{pomerleau2017fake}, the next two are inspired from \cite{wang:1}, and the last two (\textit{keyword}, \textit{proper noun overlap}) are new. 

\begin{itemize}
\item {\em N-grams match:} A \textit{n-gram} is the sequence of $n$ continuous words in a given text. The feature value is defined as the number of common n-grams in the claim and the document. 
For our system, we choose bigrams, trigrams and fourgrams. 

\item {\em Chargrams match:} Similar to n-gram, chargram is a sequence of $n$ continuous characters. 
We use bi-, tri- and four-chargrams in our system.

\item {\em Binary co-occurrence:} This feature consists of two values. The first one is the number of words of the claim that appear in the first 255 words of the document, and the second one is the number of words in the claim that appear in the entire body of the document.

\item {\em Lemma overlap:} This feature is similar to the unigram match with the difference that the words are first converted into their lemmatized form. 

\item {\em Text similarity:} We calculate the cosine similarity between the text of the claim and each sentence of the document. The maximum similarity value is considered as the feature value. 

\item {\em Word2vec similarity:} The cosine similarity between the pre-trained word2vec embeddings \cite{Mikolov:16} of the claim and the document \new{(reflecting their semantic relationship)}. 

\item {\em Keyword overlap:} 
We extract important words that appear in the text of the claim and the text of the document using \textit{cortical.io}\footnote{\url{https://www.cortical.io/}}.
The feature is defined as the number of common keywords in the claim and the document. 

\item {\em Proper noun overlap:} This feature is same as keyword overlap but instead of keywords we extract proper nouns using the NLTK Part-of-Speech tagger.\footnote{\url{https://www.nltk.org/}}
\end{itemize}

\begin{figure}
	\fbox{\includegraphics[width=0.8\textwidth]{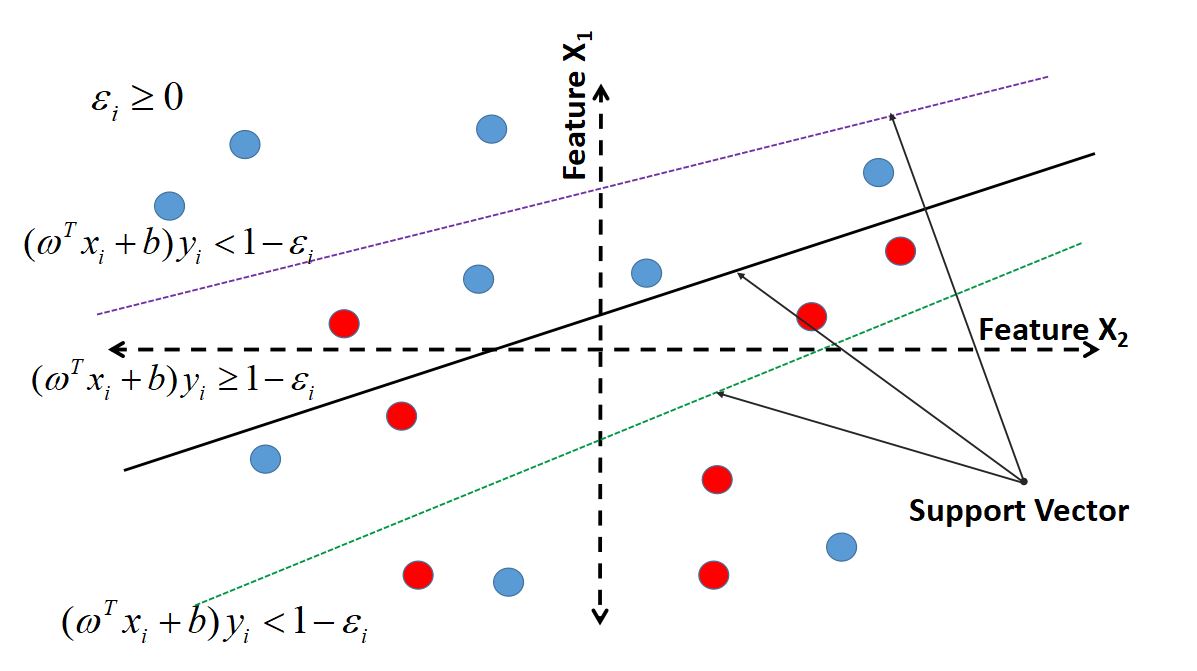}}
	\caption{Diagram outlining the SVM model used in stages 1 and 3.}
	\label{fig:stage1_3}
\end{figure}

Through cross validation, we found that a simple SVM classifier with class-wise penalty \cite{Veropoulos99controllingthe} outperforms all the other models. 
In more detail, we solve the following optimization problem:

\begin{equation}
  {  Min_{\varpi,\beta}({\frac{{\varpi}^T\varpi}{2}} + \alpha_1\sum_{i=1|\gamma_i \in Relat.}^{m} \epsilon_i +\alpha_2\sum_{i=1|\gamma_i\in Unrel.}^{m} \epsilon_i )}
  \end{equation}
Subjected to the constraints:
  \begin{equation}
      {\gamma_i({{\varpi}^T \chi_i + \beta})\geq 1- \epsilon_i} ,~~~~
      \epsilon_i\geq0 ,~~~~  i=1,...,m 
  \end{equation}
where, $\varpi$ is weight vector, $\beta$ is bias, $\gamma_i$ is the output constraint function, $\chi_i$ is the training input vector, and $\alpha_1$ and $\alpha_2$ are regularization hyperparameters of penalty terms $\epsilon_i$ for \textit{related} and \textit{unrelated} class, respectively. 
A pictorial representation of the model is depicted in Fig. \ref{fig:stage1_3}.
Hyperparameters are tuned through 10-fold cross-validation on the training dataset. 

\subsection{Stage 2: Neutral/Stance Classification}
\label{stage2}

In this stage, we require a model that is able to consider the misclassification costs of minority classes, which is vitally important for fact-checking tasks. Previous works on similar problems have shown that deep learning models \cite{ruder:8,aug:9}, supervised classifiers with sentiment-related features \cite{lai:7}, as well as sentiment features employed in neural models \cite{C18-1203}, help solving the problem effectively. We tried a variety of approaches including both neural models (Bi-LSTM, CNN) and classical machine learning models (like SVM, gradient boosting and decision trees). Based on results on a validation set, we found that a simple CNN model with embedded word vectors and sentiment features outperforms all other (and more complex) models. The top performing system of FNC-I also uses a CNN for the overall 4-class stance classification problem. 

To generate sentiment scores for the claims and documents, we use the NLTK sentiment intensity analyzer \cite{vader:13}. NLTK applies a rule-based model for sentiment analysis, and is preferred in comparison to the other systems because it is very effective in aggregating the sentiment polarity of multiple negative words. Our intuition is that a document supporting or refuting a claim will have a strong overall sentiment polarity, while a document not taking a stance towards the claim will have a more neutral overall polarity. In addition, by inspecting several documents that take a stance we noticed that the stance is usually expressed in the first few lines (this is also supported in \cite{conforti2018towards}). Thus, we consider only the document's first 10 sentences in our analysis.

\begin{figure}
	\includegraphics[width=0.99\textwidth]{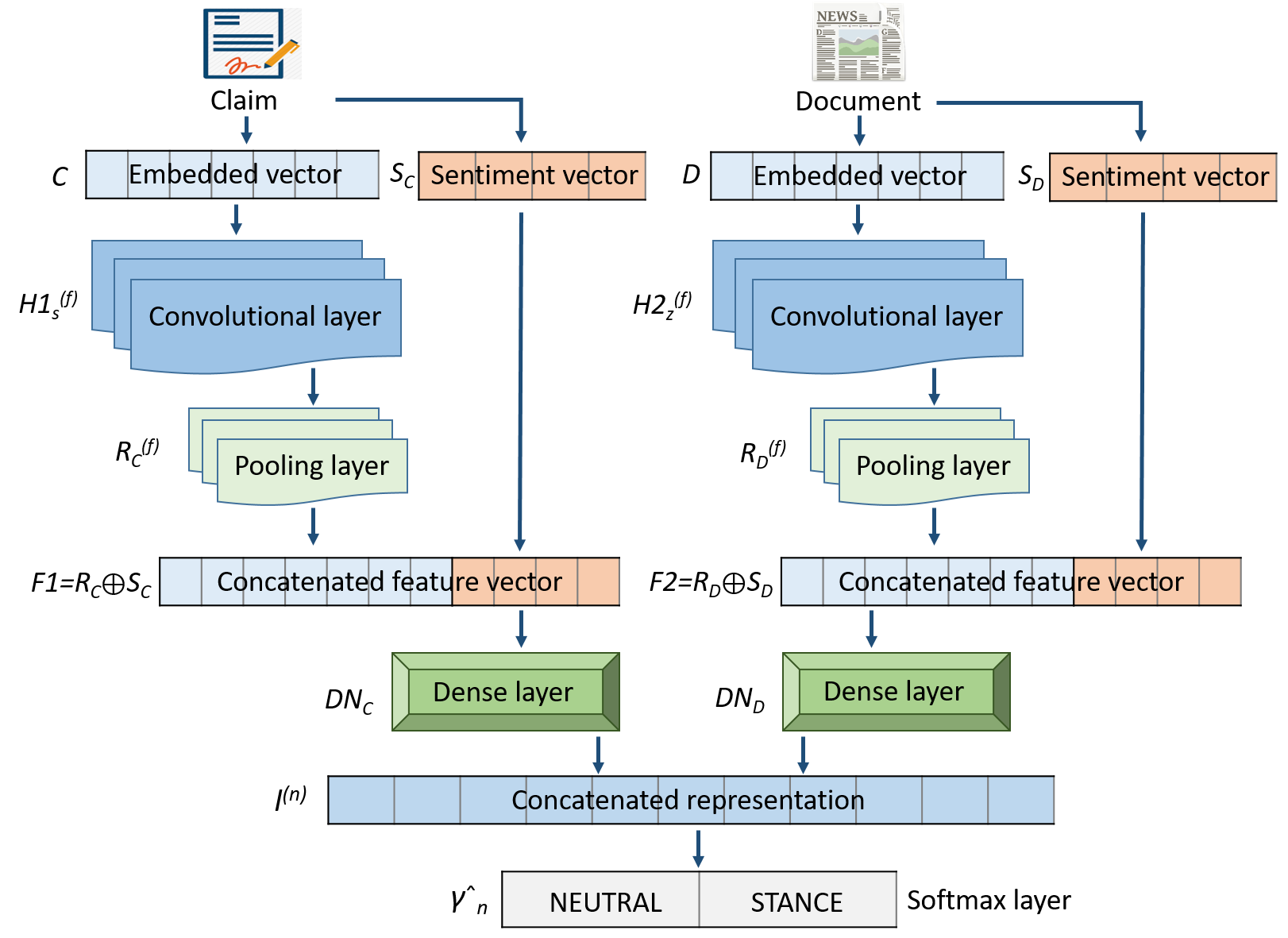}
	\caption{Diagram outlining the CNN model used in Stage 2.}
	\label{fig:stage2model}
\end{figure}

For implementing the CNN, we do not follow the approach used by the top performing system of FNC-I because the training data in our case is limited (stage 1 has filtered out \textit{unrelated} instances), and their CNN model underfits and learns only the \textit{neutral} class. Instead, we propose a simple while effective network architecture, and its  graphical representation is depicted in Fig. \ref{fig:stage2model}.  In particular, we first convert the text of the claim and the first part of the document into embedded tensors ($C$ and $D$, respectively), using the word representation method proposed by \cite{Mikolov:16}. Next, we generate an array of four sentiment scores (positive, negative, neutral, compound) for both the claim and the document using NLTK (arrays $S_C$ and $S_D$, respectively).
The two vectors $C$ and $D$ are passed through two separate convolutional networks $H1_s^{(f)}$ and $H2_z^{(f)}$, where $s$ and $z$ is the stride number in each case, each with $f$ number of $\eta \times d$ filters (where each of $f$ filters strides through $\eta$ words at a time) and stride length of $1$, i.e.:  
\begin{equation}
    H1_s^{(f)}= \phi(\sum_{i=0}^{\eta-1} \sum_{j=0}^d \varpi_{i,j}^{(f)}\cdot C_{s+i,j} + \beta^{(f)})
\end{equation}

\begin{equation}
    H2_z^{(f)}= \phi(\sum_{i=0}^{\eta-1} \sum_{l=0}^d \varpi_{i,l}^{(f)}\cdot D_{z+i,l} + \beta^{(f)})
\end{equation}
where $s=1,$...$,k-\eta+1$ ($k=$word length of claim); $z=1,$...$,p-\eta+1$ ($p=$word length of document); $\eta$ is each filter size; $d$ is dimension of embedding; $\varpi_{i,j}^{(f)}$ and $\varpi_{i,l}^{(f)}$ are weights of (neuron in position $i,j$ and $i,l$ respectively) filter $f$ connecting with $C_{s+i,j}$ and $D_{z+i,l}$ respectively; $\beta$ is bias; $f$ is filter number; and $\phi$ is a nonlinear activation function.   

These are then followed by corresponding global max-pooling layers:  
\begin{equation}
    R_C^{(f)}= max(H1^{(f)})
\end{equation}
\begin{equation}
    R_D^{(f)}= max(H2^{(f)})
\end{equation}
extracting out maximum values across each filter to obtain the network representations $R_C$ and $R_D$, respectively: 
\begin{equation}
    R_C= [R_C^{(1)}\oplus R_C^{(2)}\oplus .....\oplus R_C^{(f)}]^T
\end{equation}
\begin{equation}
    R_D= [R_D^{(1)}\oplus R_D^{(2)}\oplus .....\oplus R_D^{(f)}]^T
\end{equation}

$R_C$ is then merged with $S_C$ and $R_D$ with $S_D$ to get the final representations $ F1=R_C \oplus S_C$ and $F2 =  F2=R_D \oplus S_D$, respectively,
which are then passed through two separate multi-perceptron dense (fully connected) layers with regularization (to avoid overfitting). This gives us the two networks $DN_{C}$ and $DN_{D}$: 
 
\begin{equation}
    DN_C^{(i)}= \phi(\sum_{j=0}^{(f+4)-1}\varpi_{i,j} \cdot F1_j + \beta)
\end{equation}
\begin{equation}
    DN_D^{(i)}= \phi(\sum_{l=0}^{(f+4)-1}\varpi_{i,l} \cdot F2_l + \beta)
\end{equation}
where $f$ denotes the number of filters in the convolutional layer; $4$ is the length of $S_C$ and $S_D$; $\varpi_{i,j}$ and $\varpi_{i,l}$ are the weights of the connection between $i$th neuron of the hidden layer $DN_C$ and $DN_D$, $j$th and $l$th are the outputs of $F1$ and $F2$, respectively; $\beta$ represents bias; and $\phi$ is a nonlinear activation function. 
These layers are finally combined for the softmax binary classification: 
\begin{equation}
    I^{n}=\sum_{i=0}^{m-1} \varpi_{k,i} \cdot (DN_C^{(i)} \oplus DN_D^{(i)}) + \beta
\end{equation}
\begin{equation}
    \hat{\gamma_n}=\frac{\exp^{I^n}}{\sum_{j=0}^1\exp^{I^j}}  
\end{equation}
where $n=0,$ $1$; $\varpi_{k,i}$ is the weight of the connection between $k$th output neuron and the concatenated output of $DN_C^{(i)}$ and $DN_D^{(i)}$; and $\hat{\gamma_n}$ is the prediction of the stance of a given claim towards a given document.
As output class label we consider the one with the highest probability score. The entire network is optimised using Log likelihood cost function $L$:

\begin{equation}
   L=\sum_{n=0}^1\gamma_n \ln{\hat{\gamma_n}}
\end{equation}
where $\gamma_n$ is the true target.

\subsection{Stage 3: Agree/Disagree Classification}
\label{stage3}

Since the number of documents taking a stance is usually small, especially the number of \textit{disagree} documents, 
training a deep neural model efficiently is difficult. 
We hypothesize that a statistical machine learning algorithm trained with a well-defined set of features can be more effective. 
We again experimented with several models including SVM with and without class-wise penalty, gradient boosting, decision tree, and random forest, as well as with oversampling methods. We found that an SVM classifier similar to the one of stage 1 (depicted in Fig. \ref{fig:stage1_3}) obtains the best performance. Specifically, we effectively solve the same optimization function, while the model differs from the model of stage 1 in terms of the set of considered  features.

As mentioned before, sentiment-related features are useful in the problems related to opinion/stance classification. Thus, similar to stage 2, we again use the sentiment features $S_C$ and $S_D$ (for the claim and document, respectively) generated using NLTK. In addition, we exploit linguistic features generated using the LIWC tool (Linguistic Inquiry and Word Count).\footnote{\url{http://liwc.wpengine.com/}}
LIWC returns more than 90 features related to various linguistic properties of the input text. 
We selected the following 16 features that seem to be useful for understanding the agreement/disagreement stance: \textit{analytical thinking}, \textit{clout} (expressing confidence in perspective), \textit{authentic}, \textit{emotional tone}, \textit{conjugation}, \textit{negation}, \textit{comparison words}, \textit{affective processes}, \textit{positive emotions}, \textit{negative emotions}, \textit{anxiety}, \textit{anger}, \textit{sadness}, \textit{differentiation} (distinguishing between entities), \textit{affiliation} (references to others), and \textit{achieve} (reference to success, failure).
Moreover, we consider the \textit{refuting words} feature set used in the baseline model provided by the FNC-I organizers.
This feature set is generated by matching words from a predefined set of \textit{refuting words} with the document's words. The result is a feature vector of the length same as the number of refuting words. The vector contains \q{1} or \q{0} in each position $i$, depending on whether the  corresponding refuting word exists in the document or not.

\section{Evaluation}
\label{sec:eval}

\subsection{Evaluation Setup}
\label{sec:evalSetup}

\subsubsection{Dataset}
We use the benchmark dataset provided by the Fake News Challenge - Stage 1 (FNC-I)\footnote{\url{http://www.fakenewschallenge.org/}}  \cite{pomerleau2017fake}, which focuses on the same \textit{4-class} stance classification task (of documents towards claims). While there are datasets for related stance detection tasks available, these either address binary or three-class classification problems, or focus on detecting the stance of \textit{user opinions} regarding topics. Thus, they are not suitable to assess the 4-class stance detection problem (of Web documents) addressed in our work. 

The FNC-I dataset was derived from the Emergent dataset \cite{Ferreira2016EmergentAN} and consists of 2,587 archived documents related to 300 claims.
Each document has a summarised \textit{headline} which reflects the stance of its text (this means that each claim can be represented through different headlines of different stances which make the problem harder).
The FNC-I dataset contains 49,972 training and 25,413 test instances, related to 200 and 100 different claims, respectively.
Each instance has three attributes: a \textit{headline} (which in our case has the role of a \textit{claim}), ii) a \textit{body text} (document), and  iii) a \textit{stance label}, having one of the following values: \textit{unrelated}, \textit{discuss} (\textit{neutral}), \textit{agree}, and \textit{disagree}. 
The class distribution (Table \ref{tabdata}) shows the strong class imbalance: there is a very large number of \textit{unrelated} documents (more than 70\% in both training and testing datasets), a large number of \textit{neutral} documents (about 18\%), and a very small number of agree ($<8\%$) and disagree ($<3\%$) documents. 

\begin{table}
\caption{Data distribution of the FNC-I dataset.}
\label{tabdata}
\begin{tabular}{lrrrrr}
\toprule
\textbf{}&\textbf{All} & \textbf{Unrelated} & \textbf{Neutral} & \textbf{Agree} &\textbf{Disagree} \\
\midrule
\textbf{Train} & 49,972 & 36,545 & 8,909 & 3,678 & 840 \\
\textbf{Test}  & 25,413 & 18,349 & 4,464 & 1,903 & 697 \\
\bottomrule
\end{tabular}
\end{table}

For the first stage of our pipeline (relevance classification), we merge the classes \textit{discuss}, \textit{agree} and \textit{disagree}, to one \textit{related} class, facilitating a binary \textit{unrelated}-\textit{related} classification task. Table \ref{tabrelev} documents that \textit{unrelated} documents are more than twice the \textit{related} documents. 
For the second stage (neutral/stance classification), we 
consider only \textit{related} documents and merge the classes \textit{agree} and \textit{disagree} to one \textit{stance} class and consider the rest as \textit{neutral}. Table \ref{tabopi} shows the corresponding class distribution. We notice that the \textit{neutral} documents are about twice the \textit{stance} documents.
For the final stage (agree/disagree classification), we only consider the instances of the \textit{agree} and \textit{disagree} classes. Table \ref{tabstn} documents that the classes are imbalanced with the disagree class amounting to only 18.6\% (26.8\%) of the instances in the training (test) dataset.

\begin{table}
\caption{Class distribution for \textit{relevance} classification.}
\label{tabrelev}
\begin{tabular}{lrrr}
\toprule
\textbf{}&\textbf{Instances} & \textbf{Unrelated} & \textbf{Related} \\
\midrule
\textbf{Train} & 49,972 & 36,545 & 13,427 \\
\textbf{Test}  & 25,413 & 18,349 & 7,064  \\
\bottomrule
\end{tabular}
\end{table}

\begin{table}
\caption{Class distribution for \textit{discuss/stance} classification.}
\label{tabopi}
\begin{tabular}{lrrr}
\toprule
\textbf{}&\textbf{Instances}  & \textbf{Neutral} &\textbf{Stance}\\
\midrule
\textbf{Train} & 13,427  & 8,909 & 4,518  \\
\textbf{Test}  & 7,064   & 4,464 & 2,600  \\
\bottomrule
\end{tabular}
\end{table}

\begin{table}
\caption{Class distribution for \textit{agree/disagree} classification.}
\label{tabstn}
\begin{tabular}{lrrr}
\toprule
\textbf{}&\textbf{Instances}  & \textbf{Agree} &\textbf{Disagree} \\
\midrule
\textbf{Train} & 4,518  & 3,678 & 840 \\
\textbf{Test}  & 2,600  & 1,903 & 697 \\
\bottomrule
\end{tabular}
\end{table}

\subsubsection{Evaluation Metrics}
The FNC-I task was evaluated based on a weighted (two-level) scoring system which awards 0.25 points if a document is correctly classified as \textit{related} or \textit{unrelated}, and an additional 0.75 points if it is correctly classified as \textit{neutral}, \textit{agree} or \textit{disagree}. 
However, as argued in \cite{f1_retrospec:19}, this metric fails to take into account the highly imbalanced class distribution of the classes \textit{neutral}, \textit{agree}, \textit{disagree}. For example, a classifier that always predicts \textit{neutral} after a correct \textit{related} prediction achieves a score of 0.833, which is higher than the top-ranked system in FNC-I. 
Moreover, an effective stance classification model should perform well for the important classes, \textit{agree} and \textit{disagree}, since such documents provide crucial evidence when aiming to detect false claims or validate true claims. On the other hand, the \textit{unrelated} and \textit{neutral} classes are not important in this context. For instance, a document that discusses about a claim without taking a stance is not actually useful in fact-checking since it does not provide actual evidence about the veracity of the claim.

Based on the above observations, apart from the FNC-I evaluation measure, we also consider the following metrics for the overall assessment and comparison: i) the class-wise F1 score (the harmonic mean of precision and recall for each class), ii) the macro-averaged F1 score across all the four classes (F1$^m$), and iii) the macro-averaged F1 score across the important classes \textit{agree} and \textit{disagree} (F1$^m_{\text{Agr/Dis}}$).

\subsubsection{Baselines} 
We consider the below nine baseline methods:
\begin{itemize}
    
\item \textit{Majority vote}: The class with the maximum number of instances is always selected (\textit{unrelated} in our case).

\item \textit{FNC baseline}\footnote{\url{https://github.com/FakeNewsChallenge/fnc-1-baseline}}: A gradient boosting classifier using a set of hand-crafted features relevant for the task. The features include word/n-gram overlap features and indicator features for polarity and refutation.

\item \textit{SOLAT in the SWEN}\footnote{\url{https://github.com/Cisco-Talos/fnc-1/}} \cite{baird2017talos}: The top-ranked system of FNC-I.
This model is based on a weighted average between gradient-boosted decision trees and a deep convolutional neural network. The considered features include: word2vec embeddings, number of overlapping words, similarities between the word count, 2-grams and 3-grams, and similarities after transforming the counts with TF-IDF weighting and SVD.

\item \textit{Athene (UKP Lab)}\footnote{\url{https://github.com/hanselowski/athene_system}} \cite{hanselowski2017description}: The second-ranked system of FNC-I, based on a multilayer perceptron classifier (MLP) with six hidden and a softmax layer.
It incorporates the following hand-crafted features: unigrams, cosine similarity of word embeddings of nouns and verbs between claim and document, topic models based on non-negative matrix factorization, latent Dirichlet allocation, and latent semantic indexing, in addition to the features provided in the FNC-I baseline. 
    
\item \textit{UCL Machine Reading (UCLMR)}\footnote{\url{https://github.com/uclmr/fakenewschallenge}} \cite{riedel2017simple}: The third-ranked system of FNC-I based on simple MLP network with a single hidden layer. As features it uses the TF vectors of unigrams of the 5,000 most frequent words, and the cosine similarity of the TF-IDF vectors of the claim and document.
    
\item \textit{ComboNSE} \cite{lyon:18}: A deep MLP model which combines neural, statistical and external features. Specifically, the model uses neural embeddings from a deep recurrent model, statistical features from a weighted n-gram bag-of-words model, and hand-crafted external features (including TF, ngrams, TF-IDF, and sentiment features).
    
\item \textit{StackLSTM} \cite{f1_retrospec:19}: A model which combines hand-crafted features (selected through extensive feature analysis) with a stacked LSTM network, using 50-dimensional GloVe word embeddings \cite{pennington2014glove}, in order to generate sequences of word vectors of a claim-document pair. 

\item \textit{LearnedMMD} \cite{zhang2019stances}: A two-layer hierarchical neural network that controls the error propagation between the two layers using a Maximum Mean Discrepancy regularizer. The first layer distinguishes between the \textit{related} and \textit{unrelated} classes, and the second detects the actual stance. We report the results obtained using the provided code.\footnote{\url{https://github.com/QiangAIResearcher/hier_stance_detection}} 

\item \textit{3-Stage Trad} \cite{masood2018fake}: A three-stage classification approach similar to our pipeline method that makes use of two traditional classifiers (L1-Regularized Logistic Regression and Random Forest) and a set of 18 features. 
\end{itemize}

We compare the performance of the above baselines with our pipeline method \textit{L3S} (Learning in 3 Steps).
Whereas \cite{wang:1} may be considered as an additional baseline, we were not able to compare performance, since the data used in this method is not provided by the authors (such as the used vocabulary for contradiction indicators), and this method disregards the \textit{neutral (discuss)} class used in the FNC-I dataset and our work. 

\subsubsection{Implementation}
We use the Keras deep learning framework \cite{chollet2015keras} with TensorFlow \cite{abadi2016tensorflow} to implement our model in stage 2. For implementation of the models in stage 1 and stage 3 we use the Scikit-learn library \cite{scikit-learn}.

\subsection{Evaluation results}
\label{subsect:evalResults}
\subsubsection{Overall classification performance}

Table \ref{basecomp} shows the performance of all the approaches. 
First, we notice that, considering the problematic FNC-I evaluation measure,  \textit{CombNSE} achieves the highest performance (0.83),
followed by \textit{SOLAT}, \textit{Athene}, \textit{UCLMR}, \textit{StackLSTM}, \textit{3-Stage Trad.} (0.82) and our pipeline approach (0.81).
However, we see that the ranking of the top performing systems is very different if we consider the more robust macro-averaged F1 measure (F1$^m$).
Specifically, \textit{L3S} now achieves the highest score (0.62),  outperforming the best baseline system (\textit{stackLSTM}) by one percentage point, while \textit{CombNSE} is now in the fourth position (F1$^m$ = 0.59).

This overall performance gain of \textit{L3S} is, in particular, due to the robustness of the classifier in predicting the \textit{disagree} class, which is particularly difficult to classify due to the low number of training instances.
Specifically, our method improves the F1 score of this class by 28\% (from 0.18 to 0.23) compared to the best performing baseline for the same class (\textit{stackLSTM}). 
All other comparing baselines achieve less than $0.15$ F1 score for this class. 
With respect to the \textit{agree} class, \textit{SOLAT} is the top performing system, slightly outperforming our pipeline method by one percentage point of F1 score. However, \textit{SOLAT} performs poorly on the \textit{disagree} class achieving only 3\% F1 score. 
Considering now both \textit{agree} and \textit{disagree}, our pipeline method achieves the highest macro-averaged F1 score (F1$^m_{\text{Agr/Dis}}$), improving the state-of-the-art performance by around 12\%.

\begin{table}
\caption{Document stance classification performance.}
\label{basecomp}
\begin{tabular}{lr|r|rrrr|r}  
\toprule
System & FNC & F1$^m$ & F1$_{\text{Unrel.}}$ & F1$_{\text{Neutral}}$ & F1$_{\text{Agree}}$  & F1$_{\text{Disagr.}}$ & F1$^{m}_{\text{Agr/Dis}}$  \\
\midrule 

Majority vote  & 0.39 & 0.21 &0.84  &0.00 & 0.00  & 0.00  & 0.00\\
FNC baseline   & 0.75 & 0.45 & 0.96 & 0.69 & 0.15 & 0.02 & 0.09 \\ \midrule
SOLAT \cite{baird2017talos}         & 0.82 & 0.58 & \textbf{0.99} & 0.76 & \textbf{0.54} & 0.03 & 0.29 \\
Athene \cite{hanselowski2017description}         & 0.82 & 0.60 & \textbf{0.99} & \textbf{0.78}  & 0.49 & 0.15 & 0.32 \\
UCLMR \cite{riedel2017simple}          & 0.82 & 0.58 & \textbf{0.99} & 0.75 & 0.48 & 0.11 & 0.30  \\
CombNSE \cite{lyon:18}        & \textbf{0.83} & 0.59 & 0.98 & 0.77  & 0.49 & 0.11 & 0.30 \\
StackLSTM \cite{f1_retrospec:19}     & 0.82 & 0.61  & \textbf{0.99} & 0.76 & 0.50 & 0.18 & 0.34 \\
LearnedMMD \cite{zhang2019stances}     & 0.79 & 0.57  & 0.97 & 0.73 &0.50 &0.09  &0.29  \\
3-Stage Trad \cite{masood2018fake}     &  0.82 & 0.59   & 0.98  & 0.76   &  0.52 & 0.10  & 0.31  \\
\midrule
L3S & 0.81 & \textbf{0.62} & 0.97 & 0.75 &0.53 & \textbf{0.23} & \textbf{0.38}  \\
\bottomrule
\end{tabular}
\end{table}

With respect to the other two classes (\textit{unrelated} and \textit{neutral}), we note that  \textit{unrelated} achieves a very high F1 score for all the methods. This is expected given the nature of the classification problem and that the majority of instances belongs to this class. In the \textit{neutral} class, the top performing system (\textit{Athene}) achieves 0.78 F1 score while our approach gives 0.75. Note however that, as we have already argued, similar to the \textit{unrelated} class the \textit{neutral} class is not usually useful in a fact checking context.

\subsubsection{Detailed per-stage performance of L3S}
We now study the performance of each stage separately as well as the detailed performance of our pipeline system in terms of per-class and macro-averaged precision (P), recall (R) and F1 score (Table \ref{perform}). 

\begin{table}
\caption{Class-wise performance of the different pipeline stages and the entire pipeline system.}
\label{perform}
\begin{tabular}{lrrrr}  
\toprule
Stage & Class& P & R & F1 \\
\midrule
&Unrelated&0.97 & 0.96  & 0.97      \\
Stage 1
&Related&0.91 & 0.93  & 0.92      \\
&\textit{Macro-averaged}:  & 0.94  & 0.95 & 0.95 \\
\midrule
&Neutral&0.82 & 0.80  & 0.81      \\
Stage 2
&Stance&0.67 & 0.71  & 0.69       \\
&\textit{Macro-averaged}:  & 0.75  & 0.76 & 0.75 \\
\midrule
&Agree&0.79 & 0.75  & 0.77      \\
Stage 3
&Disagree&0.40 & 0.44  & 0.42       \\
&\textit{Macro-averaged}:&0.60 &0.60 & 0.60\\
\midrule 
&Unrelated&0.97 & 0.96  & 0.97       \\
Pipeline&Neutral&0.74     & 0.76      &0.75        \\
&Agree&0.52     & 0.53     & 0.53        \\
&Disagree&0.22      &0.23      &0.23        \\
&\textit{Macro-averaged}: &0.61  & 0.62 & 0.62\\
\bottomrule
\end{tabular}
\end{table}

With respect to stage 1 of our pipeline  (relevance classification), precision and recall of the \textit{related} class (the important class is this stage) is 0.91 and 0.93, respectively. Although the data is very imbalanced as shown in Table \ref{tabrelev} (\textit{unrelated} corresponds to around 28\% of all test instances), performance is comparably high.

Regarding stage 2 (neutral/stance classification), precision and recall of the important \textit{stance} class is 0.67 and 0.71, respectively, while that of the \textit{neutral} class is 0.82 and 0.80, respectively. In general, this task is harder than relevance classification (stage 1). Here, again the data is imbalanced, with the \textit{stance} instances corresponding to around 37\% of the test instances (see Table \ref{tabopi}). We note that there is room for improvement for the important \textit{stance} class.

The stage 3 of our pipeline deals with the harder problem of agree/disagree classification. We notice that our classifier performs well on the \textit{agree} class (P = 0.79, R = 0.75), but poorly on the \textit{disagree} class (P = 0.40, R = 0.44). We observe that, even a dedicated classifier which only considers \textit{stance} documents struggles detecting many instances of the \textit{disagree} class. Nevertheless, our method outperforms the existing methods by more than 28\% of F1 score. There are two main reasons affecting the performance of the \textit{disagree} class:
i) the classifiers of stage 2 and 3 may filter out many instances belonging to this class,
ii) the amount of training instances is limited for this class (only 840), while the data distribution is very imbalanced with the \textit{disagree} class corresponding to  18.6\% of the test instances (as shown in Table \ref{tabstn}). %
Regarding the latter, applying oversampling methods \cite{chawla2002smote} for coping with the limited amount of training instances for the minority \textit{disagree} class did not improve the performance in our experiments.

Observing precision and recall of each class for the whole pipeline approach and comparing these values with the values of the same classes in each different stage, illustrates the effects of the filtering process applied in each stage on the performance of the next stage(s). For instance, we observe that recall of the \textit{disagree} class is 0.44 in stage 3 but only 0.23 overall. The same problem exists for the \textit{agree} class (from 0.75 to 0.53). This suggests that many \textit{stance} documents are misclassified in the previous stages, making the task of stage 3 harder. We try to better understand this problem through an error analysis in the following subsection. 

\subsubsection{Error analysis}
Table \ref{confu} shows the confusion matrix of our pipeline system. 
Overall, we note that the \textit{neutral} and \textit{agree} classes seem to be frequently confused, what seems intuitive given the very similar nature of these classes, i.e. a document which discusses a claim without explicitly taking a stance is likely to agree with it. 
The results for the \textit{disagree} class illustrate that stage 1 misclassifies (as \textit{unrelated}) 18.5\% of all \textit{disagree} instances (129 instances, in total), while this percentage is less than 7\% for the \textit{agree} and \textit{neutral} classes. 
In stage 2, we see that 171 \textit{disagree} instances (25.5\%) are misclassified as \textit{neutral}, while this percentage is similar for the \textit{agree} class (26.2\%). Finally, in the last stage, 34\% of the \textit{disagree} instances are misclassified as \textit{agree}, which demonstrates the difficulty of this task. As we explained above, the highly unbalanced data distribution and the limited amount of training data are likely to contribute significantly to this picture.

\begin{table}
\caption{Confusion matrix of our pipeline system.}
\label{confu}
\begin{tabular}{lrrrrr}  
\toprule
&Agree  & Disagree &Neutral & Unrelated \\
\midrule
Agree&1,006   &278   &495   &124     \\
Disagree&237   &160   &171   &129      \\
Neutral&555   &252  &3,381   &276     \\
Unrelated&127    &31   &523 &17,668     \\
\bottomrule
\end{tabular}
\end{table}

Table \ref{confuStage} shows the confusion matrix of each stage separately, i.e., without the effect of the filtering process. We note the increasing difficulty of each stage. Stage 1 misclassifies a small number of \textit{related} instances as \textit{unrelated} (less than 8\%). Stage 2 misclassifies 29\% of the \textit{stance} instances as \textit{neutral}. Finally,  stage 3 misclassifies the majority of \textit{disagree} instances (55\%) as \textit{agree}, and around 25\% of the \textit{agree} instances as \textit{disagree}. 
It is evident from these results that there is much room for improvement for the last stage of our pipeline.

\begin{table}
\caption{Confusion matrix of different stages.}
\label{confuStage}
\begin{tabular}{lrrr}  
\toprule
&& Unrelated  & Related \\
\midrule
Stage1&Unrelated  & 17,668  & 681        \\
&Related    & 529     & 6,535        \\
\midrule
&& Neutral  & Stance \\
\midrule
Stage2&Neutral  & 3,575  & 889    \\
&Stance   & 760   & 1,840    \\
\midrule
&& Agree  & Disagree \\
\midrule
Stage3&Agree     & 1,436   & 467    \\
&Disagree  & 387  & 310    \\
\bottomrule
\end{tabular}
\end{table}

To better understand the misclassification problem, we further analyze several misclassified \textit{disagree} instances. We observe that the majority of cases concern a small number of distinct claims. In stage 1, for instance, 56/129 (43.4\%) misclassifications are associated with one claim (the claim \textit{\q{Florida woman underwent surgery to add a third breast}}). The instances of this claim were misclassified as \textit{unrelated} because of vocabulary mismatch: different words are used to express \textit{\q{breast}} in the claim and the documents (\textit{\q{boob}}, \textit{\q{breast}}), and the distance of these words in the embedded vector is unexpectedly high. 
Similarly, in stages 2 and 3, there are 40/171 (23.4\%) and 49/237 (20.7\%) misclassification cases, respectively, associated with only one claim. In stage 2 and 3, the reason is mainly due to the lack of semantic understanding of the text used to express the disagreement to the claim. For example, a misclassified document uses the phrase \textit{\q{shot down a report claiming...}} in the 2nd paragraph, while the remaining part of the document does not discuss about the claim itself. Another example of vocabulary mismatch is a misclassified document uses the text \textit{\q{all of it is bullsh*t}}, and \textit{\q{it is a nice mixture of folklore and truth}}. 
In these cases, the model fails to understand that the words \textit{bullsh*t} and \textit{folklore} negate the claim.

\section{Related Work}
\label{sec:rw}

Stance detection is a classification problem in natural language processing where the stance of a \textit{(piece of) text} towards a particular \textit{target} is explored. 
Stance detection has been applied in different contexts, including \textit{social media} (stance of a tweet towards an entity or topic) \cite{mohammad2016semeval,ijcai2017-557,aug:9,lai:7,C18-1203,ebrahimi2016weakly,xu2018cross}, \textit{online debates} (stance of a user post or argument/claim towards a controversial topic or statement) \cite{Walker:2012:SCU,P15-1012,bar2017stance,guggilla2016cnn},
and \textit{news media} (stance of an article towards a claim) \cite{pomerleau2017fake,f1_retrospec:19,lyon:18,wang:1,zhang2019stances}. Our work falls under the context of \textit{news media} where the ultimate objective is the detection of fake news. Below we discuss related works on this area and the difference of our approach.

The 4-class stance classification problem for news media was introduced in the context of the Fake News Challenge (FNC) \cite{pomerleau2017fake}, as {\em \q{a helpful first step towards identifying fake news}}.\footnote{\url{http://www.fakenewschallenge.org/}} 
The organizers made available a ground truth dataset as well as a simple baseline method that uses a set of hand-coded features and a gradient boosting classifier. 
50 teams participated in FNC using a wide array of techniques and sets of hand-crafted features. 
We consider the top-3 performing systems as baselines in our experiments (more in  Section \ref{sec:evalSetup}).

For the same problem, \cite{miller2017fake} demonstrated the challenges to apply conditional encoding and attention recurrent networks, methods known to work well in other stance detection problems.
\cite{lyon:18} proposed a deep MLP model with combined neural, statistical, and external features, achieving a state of the art performance. \cite{f1_retrospec:19} proposed the use of macro-averaged F1 score for evaluating performance in this task because this metric is less affected by highly imbalanced datasets. Moreover, the authors proposed a novel approach which uses a stacked LSTM network and 50-dimensional GloVe word embeddings \cite{pennington2014glove}, outperforming all previous methods on macro-averaged F1 score.

All these methods treat the problem as a single 4-class classification task. \cite{wang:1} proposed a different, two-stage approach where \textit{unrelated} documents are first filtered out through relevance classification, and then the \textit{related} documents are classified as \textit{contradict} (\textit{disagree}) or \textit{support} (\textit{agree}), using a gradient boosted decision tree model trained on n-gram features extracted using a specially-designed \textit{contradiction vocabulary}. However, this model ignores the \textit{discuss} class which is very prominent in fake news. Moreover, the authors do not provide the evaluation datasets and the used contradiction vocabulary. 
A more recent work \cite{zhang2019stances} also applies a two-stage approach where a first stage distinguishes \textit{related} from \textit{unrelated} documents and a second detects the actual stance (\textit{agree}, \textit{disagree}, \textit{neutral}). 
A hierarchical neural network that controls the error propagation between the two stages has been proposed. We have not been able to replicate the results as reported in \cite{zhang2019stances} after using the provided code. 
Finally, \cite{masood2018fake} proposes a three-stage approach similar to our proposed architecture that only makes use of traditional machine learning models (L1-Regularized Logistic Regression and Random Forest) and a large number of features. However, this method also fails to achieve a satisfactory performance on the important disagree class (see Section \ref{subsect:evalResults}). 

With respect to evaluation datasets, a range of datasets has been made available for related stance classification problems, for instance, for detecting the stance of ideological debates (for/against the debate topic) \cite{hasan2013stance},  context-dependent arguments/claims (pro/con a controversial statement) \cite{bar2017stance}, or tweets (favor/against a controversial topic) \cite{mohammad2016semeval}. However, the ground truth dataset provided by the FNC-I is, to the best of our knowledge, the only one focusing on the \textit{4-class} stance classification task of Web documents addressed in this paper.
Also, the aforementioned datasets  focus on detecting the stance of \textit{user opinions} regarding topics, as opposed to the stance of Web documents (like news articles) towards a given true or false claim. 

\new{
Finally, there is a number of complementary research works that are not directly concerned with stance detection but which tackle problems relevant to fact checking and fake news detection, such us trustworthiness assessment \cite{sherchan2013survey}, spammer detection \cite{bindu2018discovering,dewang2018state}, detection of check-worthy claims \cite{hassan2017toward}, or verified claim retrieval~\cite{shaar2020known}.
}

\section{Conclusion}
\label{sec:conclusion}
We have proposed a modular pipeline of cascading classifiers for the problem of document stance classification towards claims. 
This enables the use of different classifiers and features in each pipeline step, crucial to further optimize performance on a per step and class basis.
Experimental results on the benchmark dataset demonstrated the state-of-the-art performance of our pipeline model and its ability to improve the performance of the important---for fake news detection---\textit{disagree} class by 28\% (F1 score) without significantly affecting the performance of the \textit{agree} class. 
The results also showed that there is still room for further improvements, mainly due to the lack of semantic understanding of the language used to express agreement or disagreement. 
As part of future work, we are planning to focus on this problem, aiming to reduce the number of misclassified \textit{agree} and \textit{disagree} documents of each stage.

\bibliographystyle{spmpsci}    
\bibliography{bibliography}  

\end{document}